\documentclass[conference]{IEEEtran}
\IEEEoverridecommandlockouts
\usepackage{cite}
\usepackage{amsmath,amssymb,amsfonts}
\usepackage{algorithmic}
\usepackage{graphicx}
\usepackage{textcomp}
\usepackage{xcolor}
\usepackage{multirow}
\usepackage{booktabs}
\usepackage{hyperref}
\def\BibTeX{{\rm B\kern-.05em{\sc i\kern-.025em b}\kern-.08em
    T\kern-.1667em\lower.7ex\hbox{E}\kern-.125emX}}
\begin{document}


\title{Learning Stable Robot Grasping with Transformer-based Tactile Control Policies\\}

\author{En Yen Puang\textsuperscript{1}, Zechen Li\textsuperscript{2}, Chee Meng Chew\textsuperscript{2}, Shan Luo\textsuperscript{3}, Yan Wu\textsuperscript{4}

\thanks{$^{1}$E.Y. Puang is with Italian Institute of Technology and University of Genoa, Italy. {\tt\small enyen.puang@edu.unige.it}}%
\thanks{$^{2}$Z. Li and C.M. Chew are with Department of Mechanical Engineering, National University of Singapore, Singapore. {\tt\small e1192513@u.nus.edu, chewcm@nus.edu.sg}}%
\thanks{$^{3}$S. Luo is with Department of Engineering, King’s College London, United Kingdom. {\tt\small shan.luo@kcl.ac.uk}}%
\thanks{$^{4}$Y. Wu is with Institute for Infocomm Research (I$^2$R), A*STAR, Singapore. Supported by National Robotics Programme, Project M23NBK0053. {\tt\small wuy@i2r.a-star.edu.sg}.}%
}

\maketitle

\begin{abstract}
Measuring grasp stability is an important skill for dexterous robot manipulation tasks, which can be inferred from haptic information with a tactile sensor. Control policies have to detect rotational displacement and slippage from tactile feedback, and determine a re-grasp strategy in term of location and force. Classic stable grasp task only trains control policies to solve for re-grasp location with objects of fixed center of gravity. In this work, we propose a revamped version of stable grasp task that optimises both re-grasp location and gripping force for objects with unknown and moving center of gravity. We tackle this task with a model-free, end-to-end Transformer-based reinforcement learning framework. We show that our approach is able to solve both objectives after training in both simulation and in a real-world setup with zero-shot transfer. We also provide performance analysis of different models to understand the dynamics of optimizing two opposing objectives. Video results available at \url{https://stable-tactile-grasp.github.io/}.
\end{abstract}


\section{Introduction}

Tactile sensing plays an important role in robot manipulation tasks. It provides robot observations of its physical contacts and the ability for delicate interactions with the environment, which are lacking in other sensing modalities such as force-torque and vision sensors. Recent advancements and developments in vision-based tactile sensors \cite{luo2017robotic, zhang2022hardware} have cultivated wide-spread adaptation of tactile sensing in robotics applications.

With its unique advantages compared to other sensing modalities, tactile sensors are widely used in plenty of robotics applications and have become a nifty component in learning control policies. Object manipulation tasks \cite{liang2023visuo, zhang2023interaction} utilize tactile feedback in unknown environments and vision-denied scenarios. Besides, tactile sensing plays a deciding role in dexterous and dynamic manipulation tasks \cite{touch-dexterity, zhang2023visual}. Furthermore, object mapping and localization tasks \cite{luo2015localizing,bauza2019tactile, liang2021parameterized} heavily rely on tactile feedback for estimating object state that is crucial for downstream manipulation.

Robot grasping of unknown and unseen objects is another area where tactile sensing is essential. Specifically, tactile sensors are being used to estimate grasp instability caused by rotational displacement and slippage \cite{toskov2023hand, zhang2023grasp}. By detecting gravitational pivoting and slippage, tactile sensing is capable of providing necessary feedback for closed-loop control framework in achieving stable grasp. In two-finger grasping scenarios, wrench at the contact surface can be detected by tactile sensors when grasp location is far from the center of gravity of the object. Higher gripping force is able to counter the wrench resulted from the pivoting but might result in sub-optimal control policies and undesirable grasping outcomes such as causing damages to the object.

\begin{figure}[t]
  \begin{center}
    \includegraphics[width=1\linewidth]{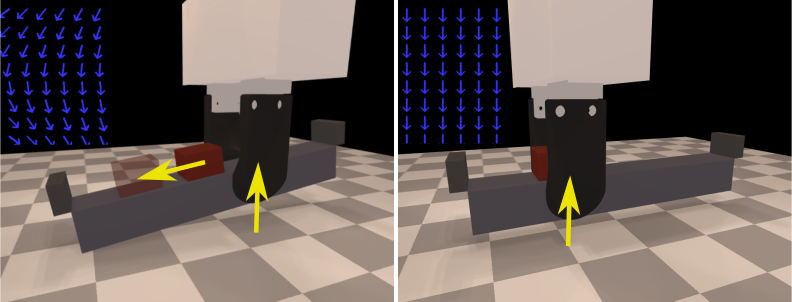}
    \vspace{-0.6cm}
  \end{center}
  \caption{Classic stable grasp task \cite{kolamuri2021improving} but with non-static load, variable weight, force control and tactile feedback (blue arrows). The sliding of load during step $t$ (left) has to be taken into account for the re-grasp in step $t+1$ (right).}
\label{fig:intro}
\end{figure}

Previous works \cite{kolamuri2021improving, li2023incipient} on tactile-based stable grasp tasks improve grasp stability by detecting rotational displacement and predicting next best grasping location using model-based algorithm or model-free Reinforcement Learning (RL) method. However, these methods do not explicitly handle objects with changing dynamics properties and hence do not optimize for gripping force. In addition, the object's weight distribution is assumed to be constant in an episode and therefore excluding objects with changing weight distributions. 

In this work, we generalize the problem and propose a new stable grasp task with 2 major modifications in the task definition:
\begin{itemize}
    \item Object's weight distribution may change as the result of every action step in an episode; and
    \item A new action dimension to control gripping force for handling wider range of weight.
\end{itemize}

The above introduced flexibility allows us to include objects with changing weight distribution when grasped. Moreover, gripping force optimization is required to handle a wider range of object's weight, improving the practicality of the task. Therefore, our main contributions of this paper are threefold: 
\begin{itemize}
    \item A new stable grasp task definition that is more challenging, inclusive and practical; 
    \item End-to-end model-free RL method for force optimized stable grasp task using tactile feedback; and
    \item Extensive analysis and benchmark on optimizing opposing objectives with different settings.
\end{itemize}

\section{Related Work}

\subsection{Handling rotational slip}
Lifting object at points far from its center of mass (CoM) often causes rotational slip. This unstable grasp scenario could lead to failure in manipulation or even mechanical issues and/or damages due to the excessive counter-torque required. \cite{castano2023measuring, toskov2023hand, li2023incipient} studied the rotation of grasped objects using tactile feedback. To improve manipulation performance, \cite{kanoulas2018center, feng2020center} study the object CoM via tactile sensors, 3D cameras and force-torque sensors. Our method only uses tactile sensor and does not explicitly handle object's rotation or CoM.

\subsection{Solving stable grasp task}
Re-grasping is a common strategy to correct grasping locations and prevent object pivoting. Previous works \cite{feng2020center, xu2023efficient, kolamuri2021improving} on stable grasping tasks focus on re-grasp planning using tactile feedback to determine object pivoting. \cite{li2023incipient, kolamuri2021improving} explicitly model object rotations using optical flows of vision-based tactile markers. Grasp location is the only degree-of-freedom in the action space. Moreover, the weight distribution of object is static in each episode. In this work, we propose an end-to-end, model-free RL method to solve a new stable grasp task. We add an action dimension to cope with the need of controlling the gripping force for object of different weights. Besides, there is no assumption of static CoM in each episode. 

\begin{figure}[t]
    \begin{center}
        \includegraphics[width=0.99\linewidth]{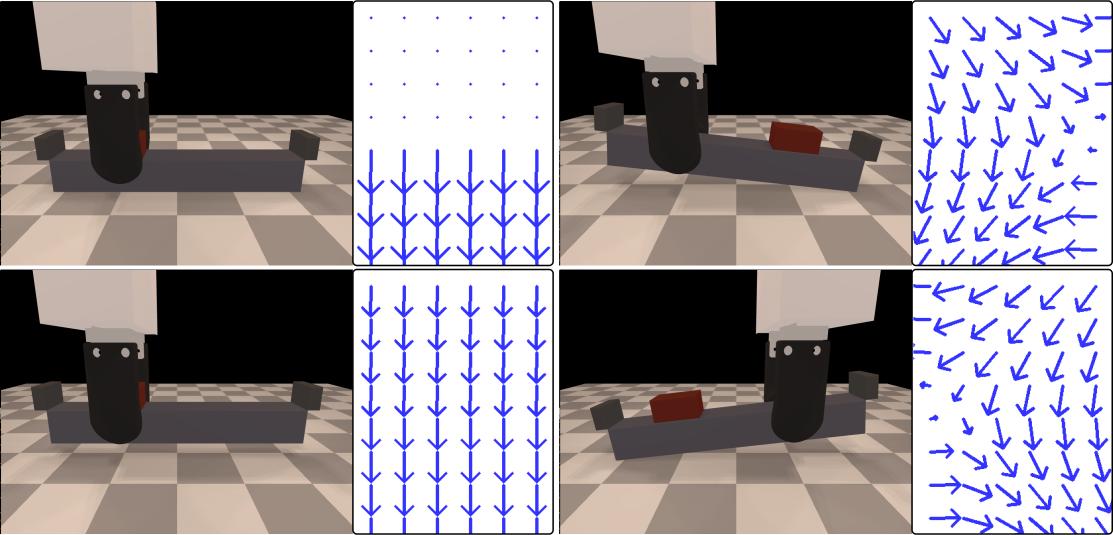}
        \vspace{-0.8cm}
    \end{center}
    \caption{Visualization of tactile maps under 4 scenarios: (top left) slippage due to insufficient gripping force; (bottom left) successful grasp at the correct location; (right) grasping at the opposite sides.}
\label{fig:tactile_scene}
\end{figure}

\subsection{Simulating tactile feedback}
We start off by running our experiments entirely in simulation. In the simulated environment, we are able to train and test our model under an extensive range of physics properties such as load density, bar friction and sensor configurations. There have been many physics engines (e.g., \textit{ODE}, \textit{Bullet}, \textit{MuJoCo}) for robotics simulation applications. In term of tactile simulation, \cite{gomes2021generation, Wang2022TACTO,gomes2023beyond,chen2023tacchi,zhao2024fots} introduce simulators for vision-based tactile sensors. These simulators use physics and rendering engines to produce high-resolution tactile images resembling the deformation on the gel surface of the optical tactile sensor. On the other hand, simulators \cite{xu2023efficient, Xu-RSS-21} simulate differentiable tactile feedback in the form of normal and shear forces. These models are suitable for low-resolution grid-based taxel and vision-based sensors with markers. We use the simulator in~\cite{xu2023efficient} to simulate two dimensional tactile shear forces, replicating the motion of a grid of markers on a vision-based tactile sensor.

\section{Methodology}

\subsection{Problem Definition}
Our stable grasp task involves grasping and lifting a rectangular bar with a two-finger parallel gripper. This is inspired by and hence similar to previous work \cite{xu2023efficient, kolamuri2021improving}, but with two major modifications. First, there is a piece of load on the bar and it is able to move along the bar. This simulates shifting center of gravity due to the tilt of the bar resulted by the previous grasp attempt. A grasp is successful when the bar, together with the load on it, is lifted without tilting. This criterion requires the control policy to observe the dynamics of the load sliding on the bar, and predicts the gripping location that is closest to the new center of gravity. Fig. \ref{fig:intro} depicts a typical episode of the task. Secondly, we add a new dimension to the action space to control the gripping force of the gripper. On top of having a successful grasp, the control policy also needs to optimize for the gripping force so that no excessive force is applied on load of various weights.

\begin{figure}[t]
    \begin{center}
        \includegraphics[width=0.99\linewidth]{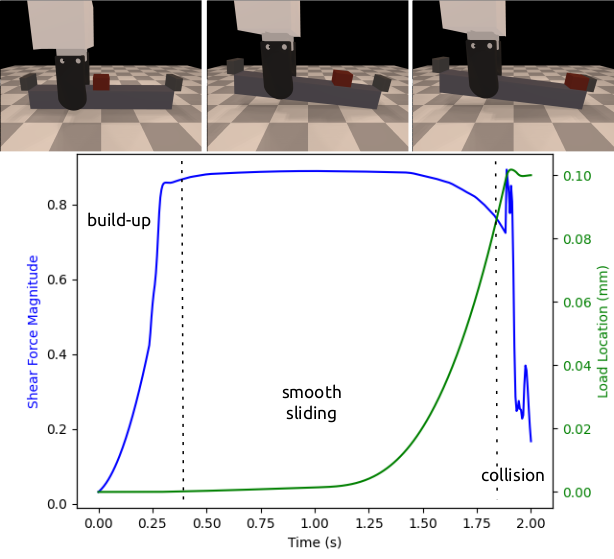}
        \vspace{-0.9cm}
    \end{center}
    \caption{Plot of average shear force magnitude of the tactile sensor (blue) over three stages when sliding occurs (green) during a grasp and lift. This shows the phases in the temporal tactile data that the control policy has to pick up.}
\label{fig:shear_mag}
\end{figure}

We frame our task as an RL task, which follows the standard \textit{act-observe} cycle. We also run our task in simulation \cite{xu2023efficient} in order to simulate a wide range of physical interactions easily before applying the learned policy directly to a number of real-world scenarios with zero-shot sim-to-real transfer. To observe the contact between the gripper and the object, we use a marker-based visuo-tactile sensor that provides shear force feedback both in simulation and in real-world. Specifically, the sensor is a 2 dimensional, low-resolution and grid-based taxels commonly seen in magnetic and vision-based models with markers. Fig. \ref{fig:tactile_scene} depicts four scenarios and their corresponding shear forces detected by the simulated tactile sensor.

\subsection{End-to-end RL framework}
\subsubsection{Features from tactile feedback}
To observe the shift of center of gravity, we treat tactile feedback as spatial-temporal input data in the RL framework. Fig. \ref{fig:shear_mag} depicts the changes in shear force magnitude at different stages when sliding occurs, from the initial build-up until collision in a low friction scenario. We sample and collect tactile maps uniformly during the lifting of the gripper. The observation space is $\Omega_o \stackrel{\text{def}}{=} \mathbb{R}^{T \times S \times F \times H \times W}$, where $T, S, F, H, W$ are the number of temporal samples, sensors, shear forces dimensions, taxel rows and taxel columns, respectively. 

\begin{figure}[t]
    \begin{center}
        \includegraphics[width=0.9\linewidth]{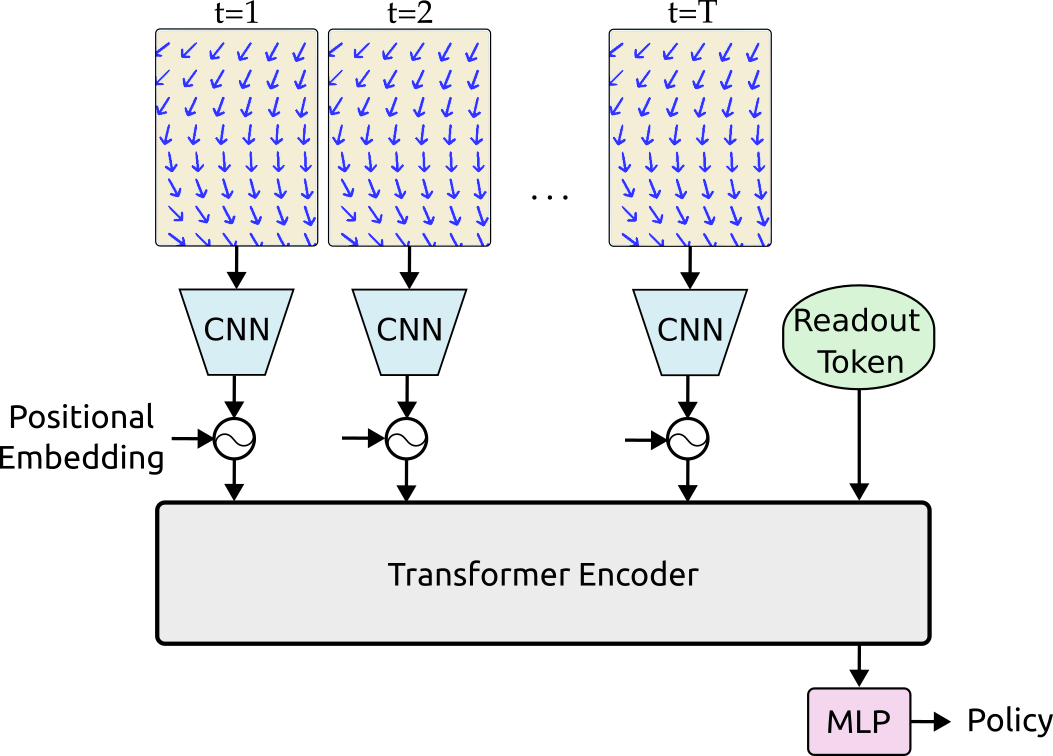}
        \vspace{-0.4cm}
    \end{center}
    \caption{Transformer-based control policy model consists of: 1. shared CNN projection block that project tactile map into token embedding; 2. Positional encoding to embed timestamp information onto tokens; 3. Learnable readout token for generating policy output.}
    \label{fig:arch_tf_a}
\end{figure}

We propose a model with a Transformer backbone \cite{vaswani2017attention} to handle tactile input in a spatial-temporal format. Specifically, our architecture is similar to ViT \cite{dosovitskiy2020image}. To get token embeddings, individual tactile map $\in \mathbb{R}^{F \times H \times W}$ first goes through a shared CNN projection block. Each token is then positionally encoded according to their timestamp. A Transformer encoder then takes in all the token embeddings from the tactile input together with an extra learnable readout token. The output embedding of this readout token is used to generate policy output through an MLP. The action space $\Omega_a \stackrel{\text{def}}{=} \mathbb{R}^2$ consists of relative changes on grasp locations on the bar and gripping force of the gripper. Fig. \ref{fig:arch_tf_a} depicts the network architecture of the transformer-based model. 

\subsubsection{Design of multi-objective reward}
The reward function for each action step $\mathcal{R}_t$ is defined as:
\begin{align}
    \mathcal{R}_t = 
    \begin{cases}
        -\text{max}\big(\delta_r,\,\delta_s\big) & \delta_r > \tau_r \text{ or } \delta_s > \tau_s \\
        \hfil \alpha(1 - \delta_f) & \hfil \text{else}
    \end{cases}
    \label{eqn:reward}
\end{align} 
where $\delta_r$ and $\delta_s \in [0, 1]$ are normalized bar tilt and bar slippage, and their corresponding thresholds $\tau$. We also normalize gripping force $f$ and load weight $w \in [0, 1]$, and get the excess force $\delta_f = f - w$. This reward function consists of two objectives: At non-terminal states, this reward function leads the control policy to stable grasp by minimizing both bar tilt and slip; At terminal states, this reward function leads the control policy to optimize gripping force by minimizing excess force $\delta_f$, given $\alpha$ as a positive scalar hyper-parameter.

\begin{figure}[t]
    \begin{center}
        \includegraphics[width=0.25\linewidth]{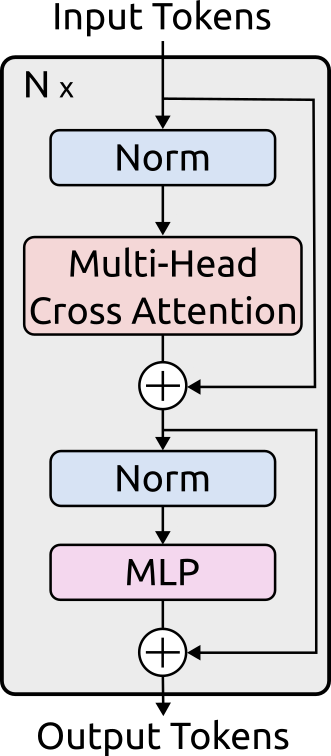}
        \vspace{-0.4cm}
    \end{center}
    \caption{Transformer-encoder is a series of $N=8$ multi-head cross attention stack. The dimension of token and MLP are 32 and 128, respectively.}
    \label{fig:arch_tf_b}
\end{figure}

\section{Experiments and Results}

\subsection{Experimental setup}

\begin{figure}[t]
    \begin{center}
        \includegraphics[width=0.9\linewidth]{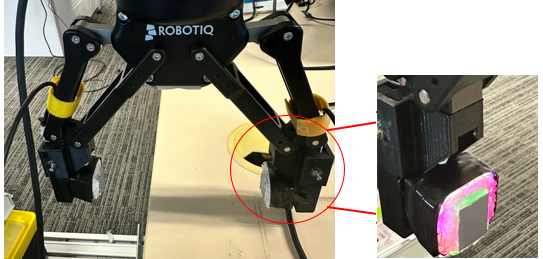}
        \includegraphics[width=0.5\linewidth]{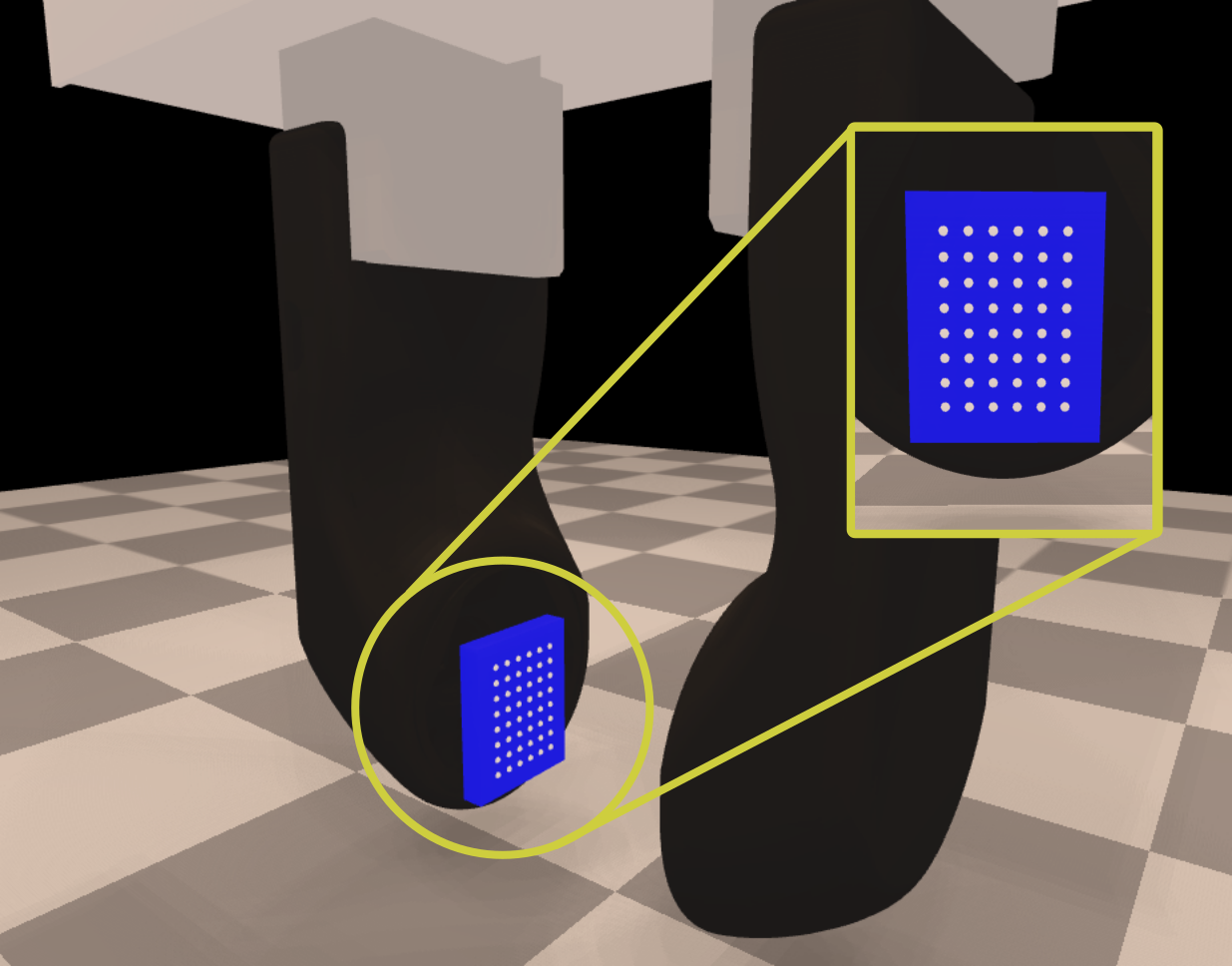}
        \vspace{-0.4cm}
    \end{center}
    \caption{Experimental setup in both simulation and real-world: Tactile sensor with 8 $\times$ 6 taxels configuration is attached to the finger of a parallel gripper. This configuration generates a tactile map of resolution 8 $\times$ 6, each bears 2 dimensional shear forces values.}
    \label{fig:setup}
\end{figure}

A tactile sensor is attached to the grasping contact surface of a parallel 2-finger gripper as depicted in Fig. \ref{fig:setup}. As opposed to having tactile readings from both fingers of the gripper, we only take observations from one (in this case, from the left finger, i.e., $S=1$). While this hardware configuration reduces the computational load, we hypothesise that it is still able to capture the necessary shear force information to solve the task. On a rectangular patch of size 14mm $\times$ 10mm, the tactile sensor has resolution 8 $\times$ 6 ($H \times W$). Each taxel bears $F=2$ dimensional shear force values, replicating 2D motion of markers on vision-based tactile sensor.

In simulation, the bar is $220mm$ long with 2 stoppers at both ends to prevent the load from falling out. Before each episode, load's density, size and starting location are randomized. Load weights $[25g, 100g]$, the bound of griping force is defined to match the bound of the load's weight (minimum force for minimum weight, and vice versa). The coefficient of friction $\mu$ between the bar and the load is randomized between 0.11 and 0.17 to cover both sliding and static scenarios. In each step, the gripper closes it fingers and lifts to a height of 20mm in 2 seconds before lowering it down. $T=11$ tactile maps are sampled uniformly during the lifting period to capture the tilt of the bar and movement of the load. A grasp is considered successful if the bar is lifted for at least 17mm ($\tau_s = 3$mm) and tilts for less than $\tau_r = 0.02$ rad. Fig. \ref{fig:intro} depicts the process of an episode.

\begin{figure}[t]
    \centering
    \begin{center}
        \includegraphics[width=\linewidth]{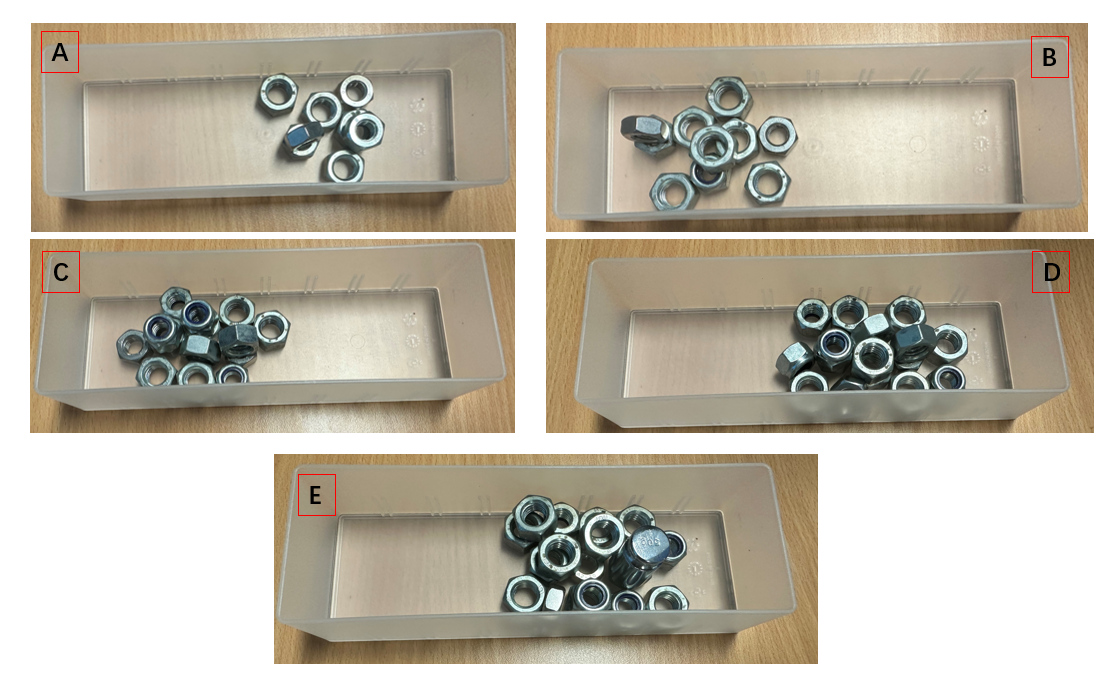}
        \vspace{-0.9cm}
    \end{center}
    \caption{The weight configurations of the real-world experimental scenarios: (A) 8 nuts; (B) 10 nuts; (C) 12 nuts; (D) 14 nuts; and (E) 15 nuts and weight.}
    \label{fig:scene_illustrate}
\end{figure}

In real-world scenarios, a Robotiq two-fingered gripper equipped with Gelsight tactile sensors, mirroring the simulation setup, is fitted on a UR5 robot to perform the stable grasping task. However, a slightly different bar construct is employed. An empty open-top plastic box (20cm x 6.5cm x 4cm) is used to carry varying numbers of nuts (10g/piece) and a 50g weight, placed randomly inside the box. A total of 5 weight configurations are used for the experiment as illustrated in Fig. \ref{fig:scene_illustrate}. For each configuration, 12 trials with random placement of the unit weights in the box are conducted, including cases with stacked weights. The combination of two different types of loads and the stacking cases covers a wider range of frictional profiles within the box, allowing for a better assessment of the zero-shot sim-to-real transferred policy's performance in real-world scenarios.

Our Transformer encoder is a series of 8 multi-head cross-attention stack depicted in Fig. \ref{fig:arch_tf_b}. Each consists of LayerNorm, Cross Attention and MLP layers. We use absolute position, sine-cosine positional embedding \cite{vaswani2017attention}. Our small transformer has token a dimension of 32, and MLP dimension of 128. The policy action head is a shallow MLP with 3 stacks of \textit{FC-ReLU}.

RL is used to train the control policy that determines the grasp location and force. The first grasp is initiated at the geometric center of the bar with half of the grasping force. Based on the tactile readings, the policy outputs the delta changes to be made in the next step on grasp location $\Delta x \in [-10mm, 10mm]$ and force $\Delta f \in [-0.125N, 0.125N]$ in order to achieve stable grasp without excessive force. Our policy is trained with an off-policy SAC \cite{haarnoja2018soft} algorithm with 8 parallel environments and 30k steps using \textit{stable-baselines3} \cite{stable-baselines3} implementation. The training takes about two and a half hours on a CPU machine.

\subsection{CNN baseline model}
\begin{figure}[ht]
    \begin{center}
        \includegraphics[width=0.55\linewidth]{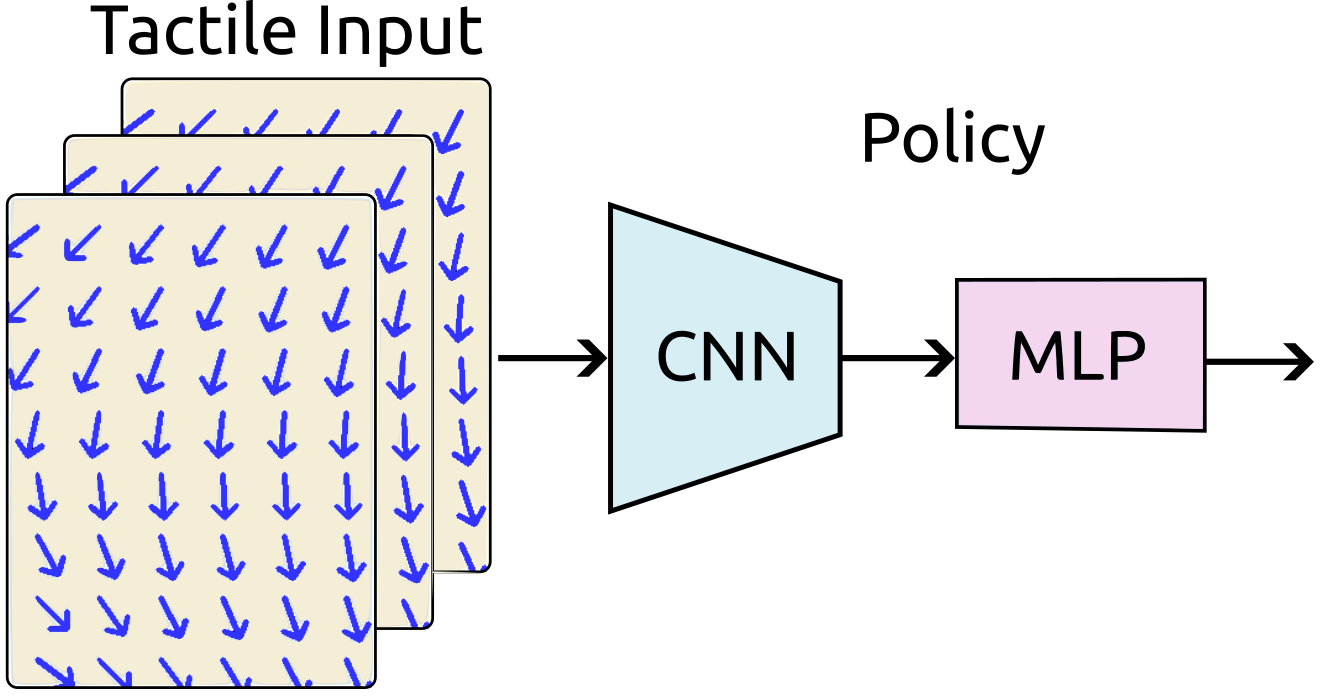}
        \vspace{-0.4cm}
    \end{center}
    \caption{We compare our Transformer-based model with a CNN-based baseline. Tactile maps sampled through time in a step are stacked channel-wise as input. It is then fed through the policy network made of CNN and MLP for action prediction.}
\label{fig:arch_cnn}
\end{figure}

We compare our method with a simple CNN-based model as baseline. To extract features from the tactile maps, we stack temporal tactile map, sensor and shear forces in the \textit{channel} dimension $\mathbb{R}^{(T \times S \times F) \times H \times W}$. The input size is effectively of $\mathbb{R}^{(11 \times 2) \times 8 \times 6}$. We use 2D convolutional layers (CNN) to extract features from spatial-temporal tactile data, followed by multi-layer perceptron (MLP) to form the control policy network. The shallow CNN block is made of 3 stacks of \textit{Conv-BatchNorm-ReLU}, while the shallow MLP is made of 3 stacks of \textit{FC-ReLU}. The MLP outputs 2 continuous action values $\mathbb{R} ^ 2$ which are then map to their corresponding physical control bounds. Fig. \ref{fig:arch_cnn} depicts the architecture of this baseline model. 

\subsection{Performance Metrics}
Success rate and average number of attempts/steps are the standard metrics used in benchmarking the stable grasp task. An successful episode has both of its bar rotation $\delta_r$ and slippage $\delta_s$ lower than their respective threshold $\tau_r$ and $\tau_s$. An episode is truncated and deemed as failed after 10 re-grasp attempts. To evaluate the optimization of gripping force, we compute the average excess force $\delta_f = f - w$ using the normalized gripping force and load weight with matching bound.

\subsection{Trade-off between minimum attempt and minimum force}
We study the trade-off between the two objectives: minimizing number of attempts and optimizing gripping force. We test the trained policies under different reward hyper-parameter $\alpha$ in simulation, each with 500 episodes and evaluate our proposed model and baseline using the aforementioned metrics, the results are shown in Tab. \ref{tab:result_tradeoff}. Our model is able to solve the task with more than 0.99 success rate, below 3 number of attempts and excess force lower than 0.2. 

Although achieving a balance between both, we found that in our experiments a policy tends to focus more on one objective, either having better success rate or lower excess force. CNN-based baseline models tend to have lower number of attempts, while Transformer-based models tend to have lower excess force. We learn that lower reward $\alpha$ improves number of attempts and the opposite lowers excess force in both types of model. This hyper-parameter intrinsically decides the focus of the policy between the opposing objectives.

\begin{figure*}[ht]
    \begin{center}
        \includegraphics[width=\linewidth]{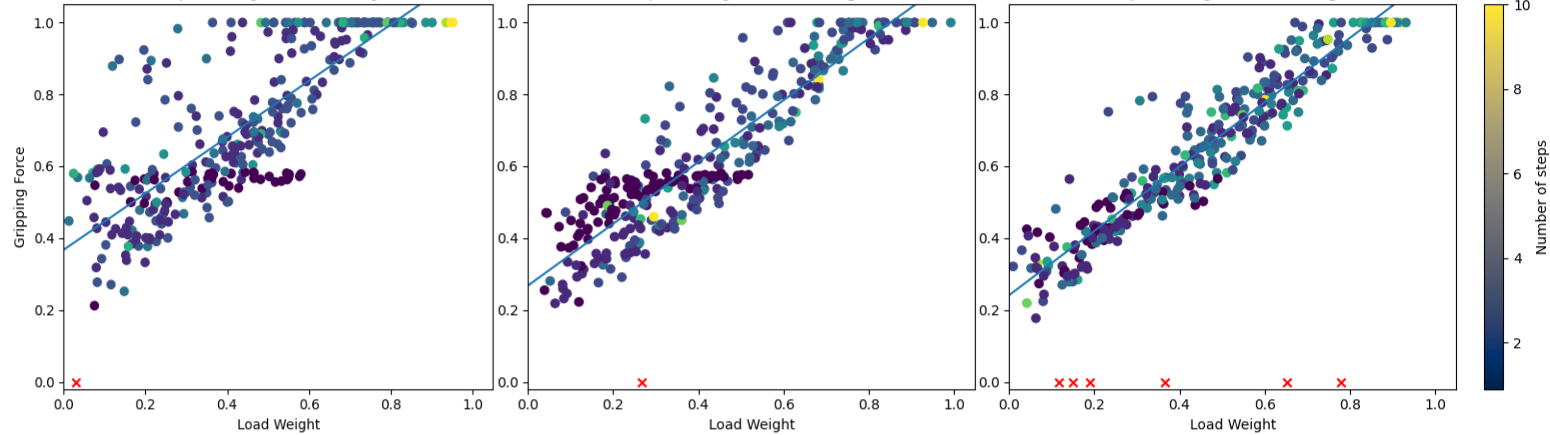}
        \includegraphics[width=\linewidth]{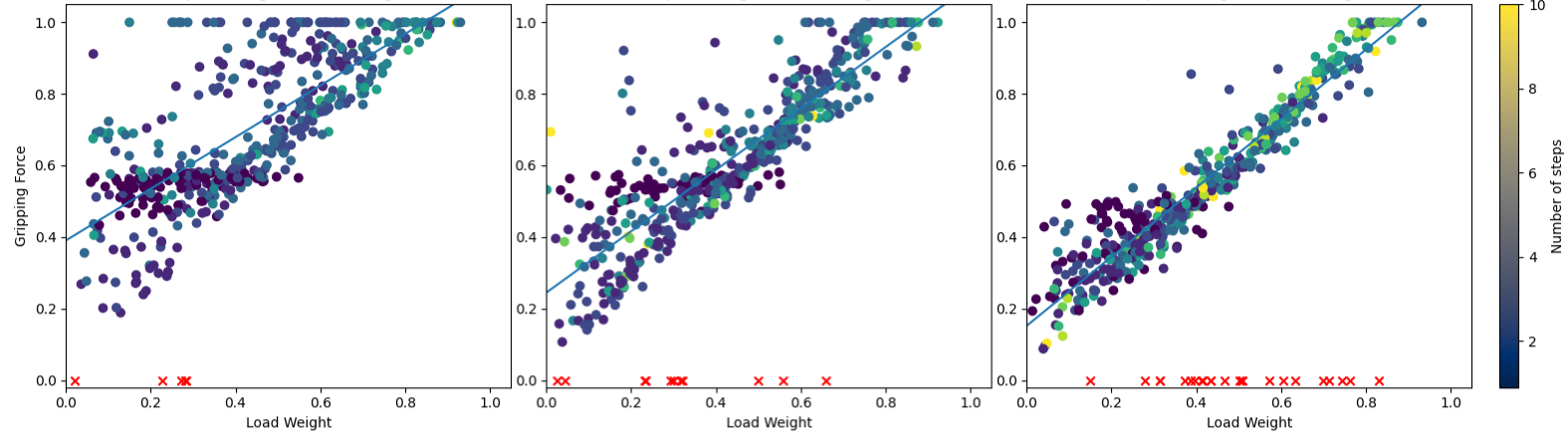}
        \vspace{-0.8cm}
    \end{center}
    \caption{CNN-based (top) and transformer-based (bottom) model's plots of gripping force against load weight over three $\alpha = [10, 30, 50]$ (left to right).}
    \label{fig:plots}
\end{figure*}

Fig. \ref{fig:plots} depicts the plots of resulted gripping force given load weight in 500 test episodes. Since the normalized gripping force and load weight are matched by design, the ideal case is when $\delta_f = 0$ and the plot is a straight line with gradient 1 and y-intercept 0. Higher $\alpha$ tends to draw the plots to the ideal line, optimizing for lower excess force. Whereas lower $\alpha$ tends to draw plots to the top left corner. We discover that failure cases (red crosses in the plots) happen more often at lower load weight range, as the policy tries to repeatedly reduce gripping force but takes up too many steps and eventually results in failure (number of attempts $>10$). 

\newcommand\Tstrut{\rule{0pt}{2.6ex}}         
\newcommand\Bstrut{\rule[-0.9ex]{0pt}{0pt}}   
\begin{table}[t]
    \centering
    \caption{Three performance metrics of the new stable grasp task. Both methods are evaluated under three different values of the reward hyper-parameter $\alpha$.}
    \begin{tabular}{cc|ccc}
	\toprule
		\multirow{2}{*}{Methods} & \multirow{2}{*}{$\alpha$} & Success & Average & Average \\ 
        && rate & \# attempts & excess force \Bstrut\\
        \midrule
        \multirow{3}{*}{CNN-based} & 10 & 0.992 & 2.887 & 0.281 \Tstrut\\
        & 30 & 0.990 & 2.762 & 0.205 \\
        & 50 & 0.986 & 3.302 & 0.194 \Bstrut\\
        \hline
        \multirow{3}{*}{tfmer-based} & 10 & 0.990 & 3.071 & 0.271 \Tstrut\\
        & 30 & 0.978 & 3.487 & 0.182 \\
        & 50 & 0.954 & 4.172 & 0.137 \\
        \bottomrule
    \end{tabular}
    \label{tab:result_tradeoff}
\end{table}

\subsection{Zero-Shot Sim-to-Real Transfer}

Based on the simulation results, we use the transformer model trained with $\alpha = 30$ to balance the trade-off between gripping force and success rate for validation in real-world scenarios. For each weight configuration, we compute the mean and standard deviation on the step lengths required to successfully grasp the bar. The experimental results are shown in Tab. \ref{tab:physical_results}.

The model succeeded in grasping the box for all 60 trials in the real-world experiment with a mean number of steps less than 3 to reach terminal state, which is reasonable given that the theoretical minimum number of steps is approximately 2 for objects with unknown fixed CoM. It is also noted that as the box is intentionally non-rigid to increase the complexity when the true CoM is nearer to the middle of the box. In this experiment, one interesting observation is that the model performs the best when the load is moderate (Scenario C). One possible explanation is that the threshold set to evaluate success matches well with the mass of the object. When the mass is too small, the discretised action may not result in observable change. When the mass is too large, the threshold may be too small to observe for even the smaller change in action.

\begin{table}[t]
    \centering
\caption{Average value and standard deviation of step length in experiment }
\label{tab:physical_results}
    \begin{tabular}{ccccccc} 
\toprule[0.6mm]
         Scenario& A & B & C & D & E & All\\ 
\midrule
         Mean & 2.429 & 2.286 & 2.267 & 2.333 & 2.813 & 2.432 \\
\midrule
         SD   & 0.821 & 0.700 & 0.573 & 0.596 & 0.882 & 0.755 \\
\bottomrule[0.6mm]
\end{tabular}
\end{table}

\section{Conclusion}
We propose a revamped version of stable grasp task that is more challenging and practical for real-world applications. This new stable grasp task features shifting center of gravity and gripping force optimization aspects, making it more inclusive in term of the type of objects the trained control policy can be applied to. We further propose a Transformer-based model and solve the task in the RL framework. We show that our model is able to solve the task with both objectives after training. We also provide analysis of the dynamics of optimizing two opposing objectives.

Transformer-based models could be further adapted for handling unevenly sampled temporal data. Positional encoding gives Transformers flexibility in embedding timestamp information into sequential data. CNN-based models are, on the other hand, more rigid in handling sequential data, especially when temporal data are simply just stacked together. This advantage makes Transformer-based models a better option for real-world applications.


\bibliography{bib}

\end{document}